\documentclass[journal]{IEEEtran}
\usepackage{amsmath}
\usepackage{amssymb}
\usepackage{bbm}
\usepackage{bm}
\usepackage{booktabs}
\usepackage{color}
\usepackage{algorithm}
\usepackage{algorithmicx}
\usepackage{algpseudocode}
\usepackage{array}
\usepackage{amsmath}
\usepackage{makecell}
\usepackage{graphicx}
\usepackage{multirow}
\usepackage{bbding}
\ifCLASSINFOpdf
\else
\fi

\begin{document}
%
\title{Multi-task Pre-training Language Model for Semantic Network Completion}


\author{\IEEEauthorblockN{ Da Li\IEEEauthorrefmark{1},
Sen Yang\IEEEauthorrefmark{2}\IEEEauthorrefmark{3},
Kele Xu\IEEEauthorrefmark{4}\IEEEauthorrefmark{5},
Ming Yi\IEEEauthorrefmark{1}, 
Yukai He\IEEEauthorrefmark{1}, and
Huaimin Wang\IEEEauthorrefmark{4}\IEEEauthorrefmark{5}}
\\
\IEEEauthorblockA{\IEEEauthorrefmark{1} Tencent, Beijing 100084, China}\\
\IEEEauthorblockA{\IEEEauthorrefmark{2} Beijing Institute of Microbiology and Epidemiology, Beijing, 100071, China}\\
\IEEEauthorblockA{\IEEEauthorrefmark{3}State Key Laboratory of Pathogen and Biosecurity, Beijing, 100071, China}\\
\IEEEauthorblockA{\IEEEauthorrefmark{4} National University of Defense Technology, Changsha, 410073, China}\\
\IEEEauthorblockA{\IEEEauthorrefmark{5} National Key Lab of Parallel and Distributed Processing, Changsha, 410073, China}

}

%



\IEEEtitleabstractindextext{%
\begin{abstract}
Semantic networks, such as the knowledge graph, can represent the knowledge leveraging the graph structure. Although the knowledge graph shows promising values in natural language processing, it suffers from incompleteness.
This paper focuses on knowledge graph completion by predicting linkage between entities, which is a fundamental yet critical task. 
Semantic matching is a potential solution for link prediction as it can deal with unseen entities, while the translational distance based methods struggle with the unseen entities. 
However, to achieve competitive performance as translational distance based methods, semantic matching based methods require large-scale datasets for the training purpose, which are typically unavailable in practical settings. 
Therefore, we employ the language model and introduce a novel knowledge graph architecture named LP-BERT, which contains two main stages: multi-task pre-training and knowledge graph fine-tuning. In the pre-training phase, three tasks are taken to drive the model to learn the relationship information from triples by predicting either entities or relations.
While in the fine-tuning phase, inspired by contrastive learning, we design a triple-style negative sampling in a batch, which greatly increases the proportion of negative sampling while keeping the training time almost unchanged.
Furthermore, we propose a new data augmentation method utilizing the inverse relationship of triples to improve the performance and robustness of the model.
To demonstrate the effectiveness of our proposed framework, we conduct extensive experiments on three widely-used knowledge graph datasets, WN18RR, FB15k-237, and UMLS. 
The experimental results demonstrate the superiority of our methods, and our approach achieves state-of-the-art results on the WN18RR and FB15k-237 datasets. 
Significantly, the Hits@10 indicator is improved by 5\% from the previous state-of-the-art result on the WN18RR dataset while reaching 100\% on the UMLS dataset.
\end{abstract}

}

\maketitle

\IEEEdisplaynontitleabstractindextext

%
\IEEEpeerreviewmaketitle

\section{Introduction}
The applications of knowledge graphs (KG) seems to be evident in both the industrial and in academic fields \cite{kazemi2018simple}, including the question answering, recommendation systems, natural language processing \cite{sundermeyer2015feedforward}, etc. These evident applications in turn have attracted considerable interest for the construction of large-scale KG. Despite the sustainable efforts that have been made, many previous knowledge graphs suffered from the incompleteness \cite{rossi2021knowledge}, as it is difficult to store all these facts in one time. To address this incompleteness issue, many link prediction approaches have been explored, with the aim to discover unannotated relations between entities to complete knowledge graphs which is challenging but critical due to its potential to boost downstream applications.

The link prediction for KG is also known as KG completion.
Previous link prediction methods can be classified into two main categories: translational distance based approach and semantic matching based approach \cite{wang2017knowledge}.
Translational distance based models typically embed both entities and relations into vector space and exploit scoring functions to measure the distance between them. Although the distance representation of entity relations can be very diverse, it is difficult to predict the entity information which did not appear in the training phase. As a promising alternative, semantic matching based approaches utilize semantic information of entities and relationships, being capable of embedding those unseen entities based on their text description. Furthermore, due to the high-complex structure of the model and the slow training speed, the proportion of negative sampling is much lower for the training purpose, which leads to insufficient learning of negative sample entity information in the entity library and severely constrains the performance of the model.

To address the aforementioned issues, especially, with the goal to alleviate poor prediction performance of the unseen node of the translation distance model and insufficient training of the text matching model, in this paper, we propose a novel pre-training framework for knowledge graph, namely LP-BERT. 
Specifically, LP-BERT employs the semantic matching representation, which leverages the multi-task pre-training strategy, including masked language model task (MLM) for context learning, masked entity model task (MEM) for entity semantic learning, and mask relation model task (MRM) for relational semantic learning. With the pre-training tasks, LP-BERT could learn relational information and unstructured semantic knowledge of structured knowledge graphs. 
Moreover, to solve the problem of insufficient training induced by the low negative sampling ratio, we propose a negative sampling in a batch inspired by contrastive learning, which significantly increases the proportion of negative sampling while ensuring that the training time remains unchanged.
At the same time, we propose a data augmentation method based on the inverse relationship of triples to increase the sample diversity, which can boost the performance further. 

To demonstrate the effectiveness of robustness of the proposed solution, we comprehensively evaluate the performance of LP-BERT on WN18RR, FB15k-237, and UMLS datasets.
Without bells and whistles, LP-BERT outperforms a group of competitive methods \cite{song2021rot,gao2021quatde,peng2020lineare,zhang2020learning,wang2021structure} and achieves state-of-the-art results on WN18RR and UMLS datasets. The Hits@10 indicator is improved by 5\% from the previous state-of-the-art result on the WN18RR dataset while reaching 100\% on the UMLS dataset.

The structure of the remainder paper is as follows. Section \ref{relatedwork} discusses the relationship between the proposed method and prior works. In Section \ref{methods}, we provide the details of our methodology, while Section \ref{exp} describes comprehensive experimental results. Section \ref{conlusion} provides the conclusion of this paper.

\section{Related Work}{\label{relatedwork}}

\subsection{Knowledge Graph Embedding}
KG embedding is a well-studied topic, and comprehensive surveys can be found in \cite{wang2017knowledge,goyal2018graph}. Traditional methods could only utilize the structural information which is observed in the triple to complete the knowledge graph. For example, TransE \cite{bordes2013translating} and TransH \cite{wang2014knowledge} were two representative works that iteratively updated the vector representation of entities by calculating the distance between entities. Convolutional Neural Network (CNN) also obtained satisfying performance in knowledge graph embedding \cite{dettmers2018convolutional,nguyen2019convolutional,schlichtkrull2018modeling}. In addition, different types of external information, such as entity types, logical rules and text descriptions, were introduced to enhance results \cite{wang2017knowledge}. For text descriptions, \cite{socher2013reasoning} firstly represented entities by averaging word embeddings contained in their names, where the word embeddings are learned from external corpora. \cite{wang2017knowledge} embedded entities and words into the same vector space by aligning the Wikipedia anchor points and entity names. CNN also has been utilized to encode word sequences in entity description \cite{zhang2016collaborative}. \cite{xiao2017ssp} proposed Semantic Space Projection (SSP) to learn topics and knowledge graph embedding together by depicting the strong correlation between fact triples and text descriptions.

Despite the satisfying performance of these models, they only learned the same text representation of entities and relationships, while the words in entity/relationship descriptions could have different meanings or importance weights in different triples. To address these problems, \cite{wang2016text} proposed a text-enhanced KG embedding model TEKE, which could assign different embeddings to relationships in different triples and use the co-occurrence of entities and words in entity-labeled text corpus. \cite{xu2016knowledge} used Long Short-Term Memory (LSTM) encoder with attention mechanism to construct context text representation with given different relationships. \cite{an2018accurate} proposed an accurate text-enhanced KG embedding method by using the mutual attention mechanism between triple specific relation mention and relation mention and entity description.

Although these methods could deal with the semantic changes of entities and relations in different triples, they could not make full use of syntactic and semantic information in large-scale free text data because they only use entity description, relation mention and word co-occurrence with entities.
KG-BERT \cite{yao2019kg}, MLMLM \cite{clouatre2020mlmlm}, StAR \cite{wang2021structure} and other works have introduced a pre-training paradigm. As aforementioned, due to the high complexities of the model and the slow training speed, the proportion of negative sampling is far lower than that of previous work, which leads to insufficient learning of negative sample entity information in the entity library. 

Compared with aforementioned methods, our method introduces the contrastive learning strategy. In the training process, a novel negative sampling approach is carried out in the batch, thus the proportion of negative sampling can be exponentially increased, which can alleviate the insufficient learning issue. Moreover, we also optimize the pre-training strategy, so that the model can not only learn context knowledge, but also learn the element information of left entity, relationship and right entity, which greatly improves the performance of the model.

\subsection{Link Prediction}
Link prediction is an active research area in KG embedding and has received lots of attention in recent years. 
KBGAT \cite{nathani2019learning}, a novel attention based feature embedding, was proposed to capture both entity and relation features in any given entity's neighbourhood. 
AutoSF \cite{zhang2020autosf} endeavored to automatically design SFs for distinct KGs by the AutoML techniques. 
CompGCN \cite{vashishth2019composition} aimed to build a novel Graph Convolutional framework that jointly embeds both nodes and relations in a relational graph, which leverages a variety of entity-relation composition operations from Knowledge Graph Embedding techniques and scales with the number of relations. 
Meta-KGR \cite{lv2019adapting} was a meta-based multi-hop reasoning method adopting meta-learning to learn effective meta parameters from high-frequency relations that could quickly adapt to few-shot relations. 
ComplEx-N3-RP \cite{chen2021relation} designed a new self-supervised training objective for multi-relational graph representation learning via simply incorporating relation prediction into the commonly used 1vsAll objective, which contains not only terms for predicting the subject and object of a given triple, but also a term for predicting the relation type. 
AttH \cite{chami2020low} was a class of hyperbolic KG embedding models that simultaneously capture hierarchical and logical patterns, which combines hyperbolic reflections and rotations with attention to model complex relational patterns.

RotatE \cite{sun2018rotate} defined each relation as a rotation from the source entity to the target entity in the complex vector space, being able to model and infer various relation patterns, including symmetry/antisymmetry, inversion, and composition. 
HAKE \cite{zhang2020learning} combined TransE and RotatE to model entities at different levels of a hierarchy while distinguishing entities at the same level. 
GAAT \cite{wang2019knowledge} integrated an attenuated attention mechanism and assigns diverse weights to relation paths to acquire the information from the neighborhoods. 
StAR \cite{wang2021structure} partitioned a triple into two asymmetric parts as in translation distance based graph embedding approach and encodes both parts into contextualized representations by a Siamese-style textual encoder. However, the pre-training stage of textual models could only learn the context knowledge of the text, while ignoring the graph structure. 

\subsection{Pre-training of Language Model}
The pre-training language model method could be divided into two categories: feature-based method and fine-tuning based method \cite{cui2021pre,qiang2021lsbert}. Traditional word embedding methods, such as Word2Vec \cite{mikolov2013distributed} and Glove \cite{pennington2014glove}, aimed to use feature-based methods to learn context semantic information and express words in the form of feature vectors. ELMo \cite{sarzynska2021detecting} extends traditional word embedding to context-aware word embedding where polysemy can be properly handled. Different from feature-based approaches, 
MLM-based pre-training method used in BERT \cite{devlin2018bert} opened up a pre-training paradigm of language model based on transformer structure. 
RoBERTa \cite{liu2019roberta} carefully measures the impact of key hyper-parameters and training data size and further enhances the effect. 
SpanBERT \cite{joshi2020spanbert} extended the BERT by masking contiguous random spans rather than random tokens, and training the span boundary representations to predict the entire content of the masked span without relying on the individual token representations within it. 
MacBERT \cite{cui2020revisiting} improves upon RoBERTa in several ways, especially the masking strategy that adopts MLM as a correction (Mac). 

Recently, the language model of pre-training has also been explored in the context of KG. \cite{wang2018dolores} learned context embeddings on entity-relational chains (sentences) generated by random walk in knowledge graph and initialized them as knowledge graph embedding models such as TransE \cite{bordes2013translating}. ERNIE \cite{zhang2019ernie} utilized knowledge graphs, which can enhance language representation with external knowledge to train an enhanced language representation model. COMET \cite{bosselut2019comet} used GPT to generate a given head phrase and a tail phrase tag of relation type in the knowledge base, which is not completely suitable for comparing the patterns of two entities with known relations. KG-BERT \cite{yao2019kg} used the MLM method for pre-training on the triple data to learn the corpus information of the knowledge graph scene.

Unlike these studies, we use a multi-task pre-training strategy based on MLM, Mask Entity Model (MEM) and Mask Relation Model (MRM) so that the model can learn not only context corpus information but also learn the association information of triples at the semantic level.

\section{Methodology}{\label{methods}}

\subsection{Task Formulation}
A knowledge graph $G$ can be represented as a set of triples $G=\{E_h, R, E_t\}_i, i\in[1, N]$, where $N$ is number of triples in $G$, $E_h$ and $E_t$ denote head entity and tail entity, and from $E_h$ to $E_t$ there existing a directed edge with attributes (i.e. relation) $R$. All entities and relations are typically short text containing several tokens. Each entity also has its description, which is a long text describing the entity in detail. We use $D_h$ and $D_t$ to denote descriptions of $E_h$ and $E_t$, respectively. The task of link prediction is to predict whether there exists a specific relation between two entities, namely, given $E_h$ and $E_t$ (with their descriptions $D_h$ and $D_t$), prediction whether $\{E_h, R, E_t\}$ holds in $G$.

\subsection{Overall Framework}

\begin{figure}[htb]
	\begin{minipage}[b]{1.0\linewidth}
		\centering
		\centerline{\includegraphics[width=9cm]{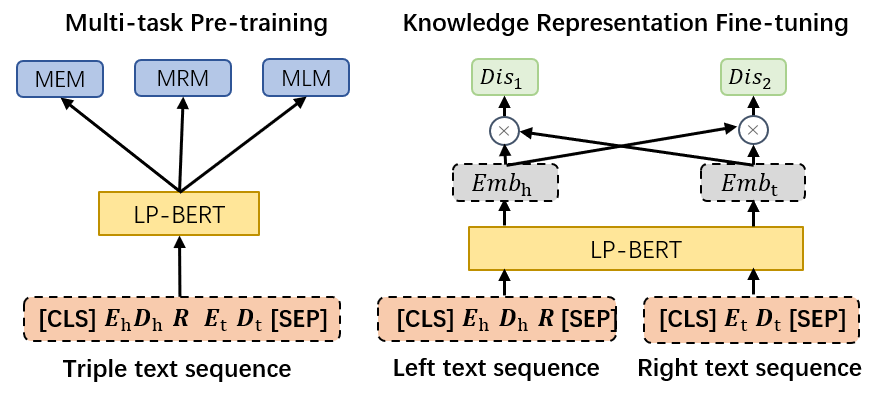}}
	\end{minipage}
	\caption{An overview of the overall framework. LP-BERT consists of two phases: pre-training and fine-tuning. In the pre-training stage, except for the Mask Language Model task (MLM), we also propose two novel pre-training tasks, which are the Mask Entity Model task (MEM) and the Mask Relation Model task (MRM). Three tasks are trained in parallel, using the multi-task manner. In the fine-tuning stage, we design a triple-style negative sampling in a batch of data, which can significantly increase the proportion of negative sampling while retaining the training time almost unchanged.}
	\label{fig:fig1}
\end{figure}

Figure \ref{fig:fig1} shows the overall framework of our proposed LP-BERT. It displays two procedures for link prediction: multi-task pre-training and knowledge fine-tuning. In the pre-training stage, in addition to the prevalent Mask Language Model task (MLM)~\cite{devlin2018bert}, we also propose two novel pre-training tasks: Mask Entity Model task (MEM) and Mask Relation Model task (MRM). Leveraging these three pre-training tasks in parallel, LP-BERT can learn both the context information of corpus and the semantic information of head-relation-tail triples.  
In the fine-tuning stage, inspired by contrastive learning, we design a triple-style negative sampling in a batch, which significantly increasing the proportion of negative sampling while keeping the training time almost unchanged. Furthermore, we propose a data augmentation method based on the inverse relationship of triples to increase sample diversity.

\subsection{Multi-task Pre-training}

\begin{figure*}[htb]
	\begin{minipage}[b]{1.0\linewidth}
		\centering
		\centerline{\includegraphics[width=18cm]{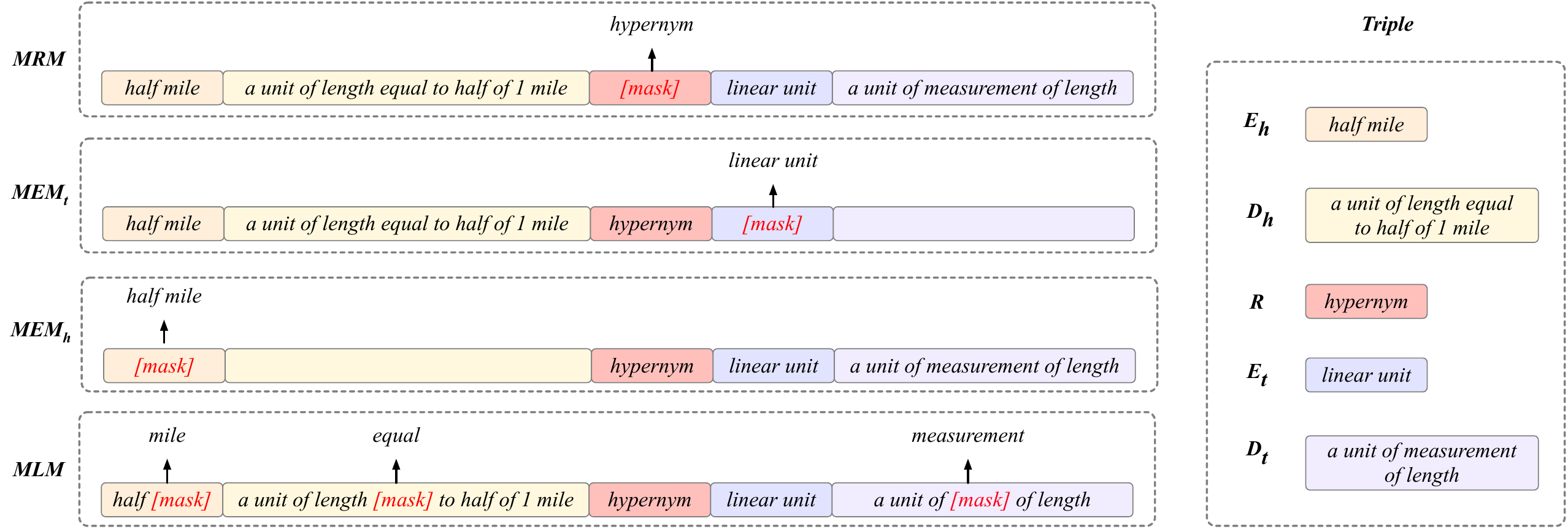}}
		\vspace{-0.3cm}
	\end{minipage}
	\caption{An instance demonstrating the pre-training tasks. Entities, relations, and entity descriptions are concatenated as a whole sequence. For MLM, random tokens are masked; For MEM, either head entity or tail is masked, so we qualify MEM with subscript ``h'' or ``t''; While For MRM, the relation is masked. It is worthwhile noticing that, these pre-training tasks can be combined using the multi-task learning paradigm during the pre-training procedure. 
	}
	\label{fig:fig2}
\end{figure*}

\begin{algorithm}
		\caption{Sample construction in the pre-training phase.}
		\label{alg.sampleconstruction}
		\begin{algorithmic}[1] 
			\Require Tokens: Token list of Triples 
			\Ensure x: masked input, $y_{1}$: MRM or MEM target, $y_{2}$: MLM target
			\Function {MLM}{$x, y_2$, region}
			\For {$i$ in region}
			\If{random[0, 1]$<$0.15}
			\State $y_2[i] = x[i]$
			\If{random[0, 1]$<$0.8}
			\State $x[i]$ = [MASK]
			\Else
			\If{random[0, 1]$>$0.5}
			\State $x[i]$ = sample(Vocab)
			\EndIf
			\EndIf
			\EndIf
			\EndFor
			\State \Return{$x, y_2$}
			\EndFunction
			\State

			\State $x\leftarrow \tilde{X}$ in Equation~\ref{eq.input}
			\State $y_1,y_2\gets$ [PAD] * len($\tilde{X}$)
			\State random\_state$\gets$ random([0, 1])

			\If{random\_state$<$0.4}
			\State $y_1[E_h]\gets x[E_h]$
			\State $x[E_h]\gets \text{[MASK]}*\text{len}(E_h)$
			\State $x[D_h]\gets \text{[PAD]}*\text{len}(E_h)$
			\State $x, y_2\gets \text{MLM}(x, y_2, R\|E_t\|D_t)$
			\EndIf
			
			\If{0.4$\leq$random\_state$<$0.8}
			\State $y_1[E_t]\gets x[E_t]$
			\State $x[E_t]\gets \text{[MASK]}*\text{len}(E_t)$
			\State $x[D_t]\gets \text{[PAD]}*\text{len}(E_t)$
			\State $x, y_2\gets \text{MLM}(x, y_2, E_h\|D_h\|R)$
			\EndIf
            
            \If{random\_state$\geq$0.8}
			\State $y_1[R]\gets x[R]$
			\State $x[R]\gets \text{[MASK]}*\text{len}(R)$
			\State $x, y_2\gets$ \text{MLM}($x, y_2$, regions of $x$ other than $R$)
			\EndIf

			\State \Return{$x, y_1, y_2$}

		\end{algorithmic}
	\end{algorithm}

We propose to pre-train the LP-BERT model with three tasks. Although they require different masking strategy, they also share the same input, which concatenates entities, relation, as well as entity descriptions in a triple as a whole sequence:
\begin{small}
\begin{align}
	\tilde{X} = [CLS] \| E_h \| D_h \| [SEP] \| R \| [SEP] \| E_t \| D_t \| [SEP]
	\label{eq.input}
\end{align}
\end{small}
Here, symbol ``$\|$'' represents sentence concatenation, ``[CLS]'' and ``[SEP]`` are two reserved tokens in BERT ~\cite{devlin2018bert}, denoting the beginning and separation of the sequence. More details of the three pre-training tasks are described in the following sections.

\subsubsection{Mask Entity Modeling (MEM)}

In the Mask Entity Modeling task, the entity sequence of the input is masked, and the model is required to recover the masked entity based on another entity and relation. The corresponding entity description is also masked to avoid information leaking. 
Since each triple includes two entities, MEM can mask either head entity or tail entity, but cannot mask both at the same time. Taking MEM on the tail entity as an instance, the input is as follows:
\begin{small}
\begin{align}
	\tilde{X} = [CLS] \| E_h \| D_h \| [SEP] \| R \| [MASK] \| [PAD] \| [SEP]
\end{align}
\end{small}

Here, we use ``[MASK]'' to represent those masked tokens. Since the tail entity is masked, its description will be replaced with reserved token ``[PAD]'' to avoid the model inferring entity just from its description. Note that both ``[MASK]'' and ``[PAD]'' could contain multiple tokens because they share the same length with the tail entity and tail description. The prediction objective is to recover the tail entity, not including its description. Similarly, for the prediction of head entities, LP-BERT just masks the head entity and predicts corresponding tokens. The head and tail entities are masked randomly with the same probabilities in the pre-training procedure.

To predict tokens in the entity, an classifier combining a multi-layer perceptron and Batch-Norm layer is built on the top of LP-BERT encoder to  output the probability matrix of the prediction results:
\begin{align}
	p=\text{BN}\circ \text{GeLU}\circ \text{MLP}\circ \text{LP-BERT}(\tilde{X})
\end{align}
in which ``$\circ$'' denotes function composition. Each token has a probability vector of the word vocabulary size, but the prediction results are not involved in the loss calculation except tokens of the masked entity. 
 
 \subsubsection{Mask Relation Modeling (MRM)}

A similar sample construction strategy as MEM is conducted for the Mask Relation Modeling task (MRM) (as shown in Algorithm 1). Instead of masking one of two entities in triple, MRM replaces tokens in the relation with ``[MASK]'', while preserving the head and tail entities and descriptions. Then MRM drives the model to predict the corresponding relation between two entities. The masked sample can be represented as follows:
\begin{align}
	\tilde{X} = [CLS] \| E_h \| D_h \| [SEP] \| [MASK] \| E_t \| D_t \| [SEP]
\end{align}

\subsubsection{Mask Language Modeling (MLM)}
In order to coexist with MEM and MRM, unlike BERT which employs \cite{devlin2018bert} random masking prediction of all tokens in the sequence, the proposed MLM method only makes local random masking for the specific text range of the sample. The random masking strategy is as follows:
\begin{itemize}
	\item For head entity prediction task in MEM, random mask only in token sequences of $E_t$ and $D_t$;
	\item For tail entity prediction task in MEM, random mask only in token sequences of $E_h$ and $D_h$;
	\item For MRM task, random mask only in token sequences of $E_h$, $D_h$, $E_t$ and $D_t$.
\end{itemize}
In this way, MLM drives the model to learn the corpus's context information. More important, though MRM and MEM are exclusive, they are both compatible with MLM. Therefore, MLM is conducted together with either MEM or MRM during the pre-training procedure for a minibatch. At the same time, doing masking is equivalent to doing dropout-like regularization for the text features of MEM and MRM tasks, and it can improve the performance of MEM and MRM, as shown in the experiments.

Algorithm~\ref{alg.sampleconstruction} shows more details about construction samples for pre-training LP-BERT. Specifically, line 20-25 and line 26-31 shows the procedure of MEM (for head entity and tail entity, respectively) with MLM, and line 32-36 shows the procedure of MRM with MLM. Line 1-15 of Algorithm~\ref{alg.sampleconstruction}  displays in detail the strategy for masking tokens, i.e., MLM.

\subsubsection{Pre-traning Loss Designing}
Since the strategies of constructing samples in MEM and MRM tasks are mutually exclusive, the prediction of head entity and tail entity cannot be simultaneously predicted for the triple samples trained by the same input model. To ensure the generalization ability of the model, we use Mask Item Model (MIM) task as a unified expression of MEM and MRM tasks and define loss function as follows:
\begin{equation}
	L = L_{MLM}(y',y)+L_{MIM}({y',y|}\alpha)
\end{equation}
where $L$ is the final loss, $y'$ and $y$ are the prediction objectives and the gold label, respectively, $\alpha$ is the random number uniformly distributed in the interval $[0,1]$. Details of $L_{MIM}$ are shown as follows:
\begin{equation}
	L_{MIM}({y',y|}\alpha)=\left\{
	\begin{array}{rcl}
		L_{MEM_h}(y',y)\;{0.0 \leq \alpha} < 0.4\\
		L_{MEM_t}(y',y)\;{0.4 \leq \alpha < 0.8}\\
		L_{MRM}(y',y)\;{0.8 \leq \alpha < 1.0}
	\end{array} \right.
\end{equation}

\subsection{Knowledge Graph Fine-tuning}

\subsubsection{Knowledge Representation}
The sample construction method in the fine-tuning stage is different from that in the pre-training stage. Inspired by the StAR \cite{wang2021structure}, for each triple, we concatenate $E_h$ and $R$ texts. Then the pre-trained LP-BERT model uses the structure of Siamese \cite{mueller2016siamese} to encode $E_h\| R$ and $E_t$, respectively. The fine-tuning objective of the model is to make the two representations of the positive sample closer and the negative further. Here, the positive sample means the ( $E_h$, $R$, $E_t$) exists in the knowledge base, while the negative sample does not.

\begin{figure}[!htbp]
	\begin{minipage}[b]{1.0\linewidth}
		\centering
		\centerline{\includegraphics[width=5cm]{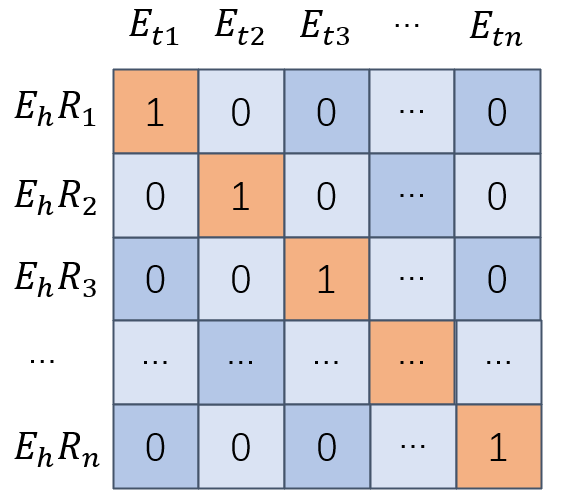}}
	\end{minipage}
	\caption{Label matrix for a batch of size $n$. For $E_hR_i$ and $E_{ti}$ in the $i$th($1\leq i\leq n$) element of batch, they can only generate 1 positive sample and $n-1$ negative samples. Therefore, there are $n^2$ samples in the batch, including $n$ positive samples and $n\dot{(n-1)}$ negative samples.}
	\label{fig:fig3}
\end{figure}

Since a triple ($E_h$, $R$, $E_t$) is splitted into $E_h\| R$ and $E_t$, the knowledge graph can only supply positive samples. Hence, to conduct binary classification during fine-tuning, we propose a simple but effective method to generate negative samples. As illustrated in Figure \ref{fig:fig3}, for a mini batch of size $n$, by interleaving $E_{h_i}\| R_i$ and $E_{t_j}$ ($1\leq i,j\leq n$), we take $E_{h_i}\| R_i, E_{t_j}$ ($i\neq j$) as negative samples. Therefore, for a mini-batch, LP-BERT only forwards twice for $n^2$ sample distance, which greatly increases the proportion of negative sampling and reduces the training time. The detailed procedure of constructing negative samples for fine-tunning is shown in Algorithm~\ref{alg.negtivesample}.

\begin{algorithm}
	\caption{Batch sample construction in the fine-tuning phase.}
	\label{alg.negtivesample}
	\begin{algorithmic}[1] 
		\Require $x_1$: $E_hR$ batch tokens, $x_2$: $E_t$ batch tokens, dict: a dict can get positive Entities for each $E_h\|R$
		\Ensure p: model predict results, y: ground truth
		\State Emb$_1$ $\leftarrow$ Encoding of $x_1$
		\State Emb$_2$ $\leftarrow$ Encoding of $x_2$ 
		\State p $\leftarrow$ cosine similarity of Emb$_1$ and Emb$_2$
		
		\State $y = []$
		\For {$E_hR \in x_1$}
		\For {$E_t\in x_2$}
		\If{$E_t\in$ dict[$E_h\|R$]}
		\State $y.append(1)$
		\Else
		\State $y.append(0)$
		\EndIf
		\EndFor
		\EndFor
		\State \Return{$p, y$}
	\end{algorithmic}
\end{algorithm}

\subsubsection{Triple Augmentation}
The above pair-based knowledge graph representation method has limitations because it cannot represent $(E_h, RE_t)$ pair directly in the head entity prediction task. Specifically, for the construction of the negative samples, we can only make negative sampling for $E_t$ but cannot for $E_h$, limiting the diversity of negative samples, especially when there are mo relationships in the dataset. 
Therefore, we propose a dual relationship data enhancement method. For each relationship $R$, we define a corresponding inverse relationship $R_{rev}$. For the head entity prediction sample in the form of $(?,R,E_r)$, we rewrite it into the form of $(E_r,R_{rev},?)$ to enhance the data. 
In this way, we can use the vector representation of $(E_tR_{rev}, E_h)$ to replace $(E_h, RE_t)$, improving the diversity of sampling and the robustness of the model. Moreover, we use the mixed precision strategy of fp16 and fp32 to reduce the GPU memory usage of gradient calculation to improve the size of $n$ and increase the negative sampling ratio.

\subsubsection{Fine-tuning Loss Designing}

We designed two distance calculation methods to jointly calculate the loss function,
\begin{equation}
	L = L_1(V_{E_hR}, V_{E_t})+ L_2(V_{E_hR}, V_{E_t})
\end{equation}
where $V_{E_hR}$ and $V_{E_t}$ are encoded vectors of $E_hR$ and $E_t$, respectively.
\begin{small}
\begin{eqnarray}
	L_1(d_1)=\left\{
	\begin{array}{rcl}
		-\alpha_t(1-d_1)^\gamma log(d_1)	& y=1  \\
		-\alpha_t(d_1)^\gamma log(1-d_1)	& y\ne1\\
	\end{array}
	\right. \label{con:eq11}
\end{eqnarray}
\begin{eqnarray}
	L_2(d_2)=\left\{
	\begin{array}{rcl}
		Sigmoid(sum(d_2))	& y=1  \\
		1 - Sigmoid(sum(d_2))	& y\ne1\\
	\end{array}
	\right.
\end{eqnarray}
\end{small}
where
\begin{eqnarray}
	\left\{
	\begin{array}{rcl}
		d_1 = \frac{V_{E_hR}\dot{V_{E_t}}}{\Vert V_{E_hR}\Vert\Vert V_{E_t} \Vert} \\
		d_2 = \Vert V_{E_hR} - V_{E_t} \Vert
	\end{array}
\right.
\end{eqnarray}
where $\alpha$ is used to adjust the weights of positive and negative samples, $\gamma$ is used to adjust the weights of samples that are difficult to distinguish. Two different dimensional distance calculation methods are used to calculate the distance relationship between multi-task learning vector pairs.

\section{Experiments}{\label{exp}}

In this part, we firstly detail the experimental settings. Then, we demonstrate the effectiveness of the proposed model on multiple widely-used datasets, including WN18RR \cite{dettmers2018convolutional} FB15k-237 \cite{toutanova2015representing} and UMLS \cite{dettmers2018convolutional} benchmark datasets. Secondly, ablation studies are conducted and we tease apart them from our standard model and keep other structures as they are, with the goal to justify the contribution of each improvement. Finally, we conduct extensive analysis of the prediction performance for the unseen entities and a case study is also provided to show the effectiveness of proposed LP-BERT.

\subsection{Experiment Settings and Datasets}

We evaluate the proposed models on WN18RR \cite{dettmers2018convolutional}, FB15k-237 \cite{toutanova2015representing} and UMLS \cite{dettmers2018convolutional} datasets. For WN18RR, the dataset was adopted from WordNet, for the link prediction task \cite{miller1998wordnet} and it consists of English phrases and the corresponding semantic relations. FB15k-237 \cite{toutanova2015representing} is a subset of Freebase \cite{bollacker2008freebase}, which consists of real-world named entities and their relations. Both WN18RR and FB15k-237 are updated from WN18 and FB15k \cite{bordes2013translating} respectively by removing inverse relations and data leakage, which is one of the most popular benchmark. UMLS \cite{dettmers2018convolutional} is a small KG containing medical semantic entities and their relations. The summary statistics of the datasets are shown in Table \ref{tab:test4}.

\begin{table}[htbp]
	\small
	\vspace{-0.3cm}
				\centering
		\caption{\label{tab:test4} The summary statistics of the used datasets, which including WN18RR, FB15k-237 and UMLS.}
		\begin{tabular}{l|rrrr}

			\toprule
			 & WN18RR & FB15k-237 & UMLS \\
			\midrule
			Entities & 40943 & 14541 & 135\\
			Relations & 11 & 237 & 46\\
			Train samples & 86835 & 272115 & 5216 \\
			Valid samples & 3034 & 17535 & 652 \\
			Test samples & 3034 & 20466 & 661 \\
			\bottomrule
	\end{tabular}

	\vspace{-0.3cm}
\end{table}

\begin{table*}[htbp]
	\caption{\label{tab:test1}Experimental results on WN18RR, FB15k-237 and UMLS datasets. The bold numbers denote the best results in each genre while the underlined ones are state-of-the-art performance. We can see that LP-BERT achieves state-of-the-art performance in multiple evaluation results on WN18RR and UMLS datasets, and outperforms other semantic matching models on the FB15k-237 dataset. $\uparrow$ means that higher values provide better performance, while $\downarrow$ means that lower values provide better performance.}
	\centering
	\setlength{\tabcolsep}{0.6mm}{
		\begin{tabular}{l*{13}{c}}
			\toprule
			\multirow{2}{*}{\textbf {Methods}} & \multicolumn{5}{c}{\textbf {WN18RR}} & \multicolumn{5}{c}{\textbf {FB15k-237}} & \multicolumn{2}{c}{\textbf {UMLS}} & \\
			\cmidrule(l){2-6}	\cmidrule(l){7-11}	\cmidrule(l){12-13}
			& Hits@1$\uparrow$ & Hits@3$\uparrow$ & Hits@10$\uparrow$ & MR$\downarrow$ & MRR$\uparrow$ & Hits@1$\uparrow$ & Hits@3$\uparrow$ & Hits@10$\uparrow$ & MR$\downarrow$ & MRR$\uparrow$ & Hits@10$\uparrow$ & MR$\downarrow$ &\\
			\midrule
			\multicolumn{11}{l}{\textbf{\emph{Translational distance models}}} \\
			\midrule
			TransE\cite{bordes2013translating} & 0.043 & 0.441 & 0.532 & 2300 & 0.243 & 0.198 & 0.376 & 0.441 & 323 & 0.279 & 0.989 & 1.84 & \\
			DistMult\cite{yang2014embedding} & 0.412 & 0.470 & 0.504 & 7000 & 0.444 & 0.199 & 0.301 & 0.446 & 512 & 0.281 & 0.846 & 5.52 &\\
			ComplEx\cite{trouillon2016complex} & 0.409 & 0.469 & 0.530 & 7882 & 0.449 & 0.194 & 0.297 & 0.450 & 546 & 0.278 & 0.967 & 2.59 & \\
			R-GCN \cite{schlichtkrull2018modeling} & 0.080 & 0.137 & 0.207 & 6700 & 0.123 & 0.100 & 0.181 & 0.300 & 600 & 0.164 & - & -  &\\
			ConvE\cite{dettmers2018convolutional} & 0.419 & 0.470 & 0.531 & 4464 & 0.456 & 0.225 & 0.341 & 0.497 & 245 & 0.312 & \textbf{0.990} & \textbf{1.51} & \\
			KBAT\cite{nathani2019learning} & - & - & 0.554 & 1921 & 0.412 & - & - & 0.331 & 270 & 0.157 & - & -  &\\
			QuatE\cite{zhang2019quaternion} & 0.436 & 0.500 & 0.564 & 3472 & 0.481 & 0.221 & 0.342 & 0.495 & 176 & 0.311 & - & -  &\\
			RotatE\cite{sun2018rotate} & 0.428 & 0.492 & 0.571 & 3340 & 0.476 & 0.241 & 0.375 & 0.533 & 177 & 0.338 & - & -  &\\
			TuckER\cite{balavzevic2019tucker} & 0.443 & 0.482 & 0.526 & - & 0.470 & 0.266 & 0.394 & 0.544 & - & 0.358 & - & -  &\\
			AttH\cite{chami2020low} & 0.443 & 0.499 & 0.573 & - & 0.486 & 0.252 & 0.384 & 0.540 & - & 0.348 & - & -  &\\ 
			ConE\cite{bai2021modeling} & 0.453 & 0.515 & 0.579 & - & 0.496 & 0.247 & 0.381 & 0.54	& - & 0.345 & - & -  &\\
			DensE\cite{lu2020dense} & 0.443 & 0.508 & 0.579 & 3052 & 0.491 & 0.256 & 0.384 & 0.535 & 169 & 0.349 & - & - &\\
			Rot-Pro\cite{song2021rot} & 0.397 & 0.482 & 0.577 & - & 0.457 & 0.246 & 0.383	& 0.540	& - & 0.344 & - & - &\\
			QuatDE\cite{gao2021quatde} & 0.438 & 0.509 & \textbf{0.586} & 1977 & 0.489 & 0.268 & \textbf{\underline{0.400}} & \textbf{\underline{0.563}} & \textbf{\underline{90}} & 0.365 & - & -  &\\
			LineaRE \cite{peng2020lineare} & \textbf{\underline{0.453}} & 0.509 & 0.578 & 1644 & 0.495 & 0.264	& 0.391	& 0.545	& 155 & 0.357 & - & - & \\
			CapsE\cite{vu2019capsule} & - & - & 0.559 & \textbf{718} & 0.415 & - & - & 0.356 & 403 & 0.150 & - & - &\\
			RESCAL-DURA \cite{zhang2020duality} & 0.455 & - & 0.577 & - & \textbf{\underline{0.498}} & \textbf{\underline{0.276}} & - & 0.550 & - & \textbf{\underline{0.368}} & - & - &\\
			HAKE\cite{zhang2020learning} & 0.452 & \textbf{0.516} & 0.582 & - & 0.497 & 0.250 & 0.381 & - & - & 0.346 & - & - & \\
			\midrule
			\multicolumn{11}{l}{\textbf{\emph{Semantic matching models}}} &&& \textbf{\emph{Parameters}} \\
			\midrule
			KG-BERT\cite{yao2019kg} & 0.041 & 0.302 & 0.524 & 97 & 0.216 & - & - & 0.420 & 153 & - & 0.990 & 1.47 & 102M\\
			StAR\cite{wang2021structure} & 0.243 & 0.491 & 0.709 & \textbf{\underline{51}} & 0.401 & 0.205 & 0.322 & 0.482 & \textbf{117} & 0.296 & 0.991 & 1.49 & 335M \\
			\textbf{LP-BERT} & \textbf{0.343} & \textbf{\underline{0.563}} & \textbf{\underline{0.752}} & 92 & \textbf{0.482} & \textbf{0.223} & \textbf{0.336} & \textbf{0.490} & 154 & \textbf{0.310} & \textbf{\underline{1.000}} & \textbf{\underline{1.18}} & 102M \\
			\bottomrule
	\end{tabular}}
\end{table*}

We implement the LP-BERT using PyTorch \footnote{https://pytorch.org/} framework, on a workstation with an Intel Xeon processor with a 64GB of RAM and a Nvidia P40 GPU for the training purpose. The AdamW optimizer is used with $5\%$ steps of warmup. For the hyper-parameters in LP-BERT, we set the epochs to $50$, the batch size as $32$, and the learning rate=$10^{-4}/5\times10^{-5}$ respectively for the linear and attention parts initialized. The early-stop epoch number is set as $3$. In the knowledge graph fine-tuning phase, we set the batch size as $64$ on WN18RR, $120$ for FB15k-237, $128$ on UMLS based on the best Hits$@10$ on development dataset. The learning rate is set as $10^{-3}/5\times10^{-5}$ respectively for the linear and attention parts initialized with LB-BERT. The number of training epochs is $7$ on WN18RR and FB15k-237, while $30$ for the UMLS datasets, $\alpha=0.8$ on WN18RR and UMLS, $\alpha=0.5$ on FB15k-237, and $\gamma=2$ in Eq \ref{con:eq11}.

In the inference phase, all other entities in the knowledge graph are regarded as the wrong candidate which can damages their head or tail entities. The trained model aims to use the ``filtered'' settings to correct triple ranking of the corrupt. The evaluation metrics has two aspects : (1) Hits@$N$ represents the ratio of test instances in the top $N$ of correct candidates; (2) The average rank (MR) and the average reciprocal rank (MRR) reflect the absolute ranking.

\subsection{Results}

We benchmark the link prediction tasks using proposed methods and competitive approaches, including both the translational distance-based approaches and semantic matching-based methods. For the translational distance models, 18 widely-used solutions are tested in our experiments, including TransE \cite{bordes2013translating}, DistMult \cite{yang2014embedding}, ComplEx \cite{trouillon2016complex}, R-GCN \cite{schlichtkrull2018modeling}, ConvE \cite{dettmers2018convolutional}, KBAT \cite{nathani2019learning}, QuatE \cite{zhang2019quaternion}, RotatE \cite{sun2018rotate}, TuckER \cite{balavzevic2019tucker}, AttH \cite{chami2020low}, DensE \cite{lu2020dense}, Rot-Pro \cite{song2021rot}, QuatDE \cite{gao2021quatde}, LineaRE \cite{peng2020lineare}, CapsE \cite{vu2019capsule}, RESCAL-DURA \cite{zhang2020duality} and HAKE \cite{zhang2020learning}. As expected, only a few previous attempts employed semantic-matching-based methods, due to the difficulties of the training. Here, KG-BERT \cite{yao2019kg} and StAR \cite{wang2021structure} are used for the quantitative comparison. 

The detailed results are shown in Table \ref{tab:test1}. As can be observed from the Table, the LB-BERT is able to achieve state-of-the-art or competitive performance on all three widely used datasets, including WN18RR, FB15k-237 and UMLS dataset. The improvement is especially significant in terms of Hits@10 and Hits@3 due to the superior generalization performance of multi-task pre-training textual encoding approach, which will be further analyzed in the section below. Furthermore, LP-BERT surpasses all other methods by a large margin in terms of Hits@3, Hits@10 on WN18RR and Hits@10 on UMLS. Although it only achieves inferior performance on FB15k-237 dataset and Hits@1 and MRR of WN18RR dataset compared to translational distance models, it still remarkably outperforms other semantic matching models such as KG-BERT and StAR from the same genre by introducing structured knowledge. In particular, LB-BERT outperforms StAR \cite{wang2021structure}, which is the previous state-of-the-art model, on all three datasets with only one-third parameters numbers of it. For the WN18RR datasets, the Hits@1 is increased from 0.243 to 0.343, Hits@3 is increased to 0.563 from 0.491.

The experimental results show that the semantic matching models perform well in the topK recall evaluation methods, but the Hits@1 result is significantly inferior to the translation distance models. This is because that the features of the semantic matching models are based on text, which leads to the vector representation of similar entities in the text being close and difficult to distinguish. Although translation distance models perform well on the Hits@1, they are incapable of understanding the text semantics. For new entities not seen in the training set, the prediction results of translation distance models are random, while the semantic matching models are reliable, which is why Hits@3 and Hits@10 LP-BERT can far exceed the translation distance models to achieve state-of-art performance.

\begin{table*}[htbp]
	\centering
		\caption{\label{tab:test2} Quantitative comparisons with KG-BERT and StAR on WN18RR datasets. ``Train'' denote the time for per training epoch, and and ``Inference'' denotes total inference time on test set. The values were collected using Tesla P40 without mixed precision.}
	\setlength{\tabcolsep}{0.4mm}{
		\begin{tabular}{lcccccccc}
			\toprule
			& Weight Initialization & Hits@1$\uparrow$  & Hits@3$\uparrow$  & Hits@10$\uparrow$ & MR$\downarrow$ & MRR$\uparrow$  & Train$\downarrow$ & Inference$\downarrow$ \\
			\midrule
			KG-BERT & RoBERTa-base & 0.130 & 0.320 & 0.636 & 84 & 0.278 & 4h & 32h \\
			StAR & RoBERTa-base & 0.202 & 0.410 & 0.621 & \underline{71} & 0.343 & 2h & 0.9h \\
			\textbf{\emph{LP-BERT}} & \textbf{\emph{RoBERTa-base}} & \underline{0.278} & \underline{0.502} & \underline{0.708} & 79 & \underline{0.424} & 0.8h & 0.8h \\
			\midrule
			KG-BERT & RoBERTa-large & 0.119 & 0.387 & 0.698 & 95 & 0.297 & 7.9h & 92h \\
			StAR & RoBERTa-large & 0.243 & 0.491 & 0.709 & \textbf{\underline{51}} & 0.401 & 5.5h & 1h \\
			\textbf{\emph{LP-BERT}} & \textbf{\emph{RoBERTa-large}} & \underline{0.306} & \underline{0.517} & \underline{0.718} & 69 & \underline{0.444} & 2.2h & 1h \\
			\midrule
			\textbf{LP-BERT} & \textbf{LP-BERT-base} & \textbf{\underline{0.343}} & \textbf{\underline{0.563}} & \textbf{\underline{0.752}} & 92 & \textbf{\underline{0.482}} & 0.8h & 0.8h \\
			\bottomrule
	\end{tabular}}

\end{table*}

Similar to KG-BERT and StAR, our model is relied BERT and we compare LP-BERT with KG-BERT and StAR on WN18RR in detail, including different initialization manners. As shown in Table \ref{tab:test2}, LP-BERT consistently achieves superior performance over most metrics. The evaluation effect of the LB-BERT model based on RoBERTa-base has exceeded the evaluation effect of KG-BERT and StAR based on RoBERTa-large. As for empirical efficiency, due to the small number of model parameters and the strategy of negative sampling based on the batch in the training process, our model is faster than KG-BERT and StAR for both the training and inference phases.

\subsection{Ablation Study}

\begin{table*}[htbp]
	\centering
	\caption{\label{tab:test3} Ablation study for LP-BERT on the WN18RR dataset.}
	\setlength{\tabcolsep}{0.7mm}{
		\begin{tabular}{lcccccccc}
			\toprule
			MLM& MEM &MRM  & Batch-Sampling & Hits@1$\uparrow$ & Hits@3$\uparrow$ & Hits@10$\uparrow$ & MR$\downarrow$ & MRR$\uparrow$ \\
			\midrule
			 \XSolidBold & \XSolidBold & \XSolidBold & \XSolidBold & 0.136 & 0.306 & 0.499 & 143 & 0.257 \\
			 \XSolidBold & \XSolidBold & \XSolidBold & \CheckmarkBold & 0.278 & 0.502 & 0.708 & \textbf{79} & 0.424 \\
			 \CheckmarkBold & \XSolidBold & \XSolidBold & \CheckmarkBold &  0.307 & 0.530 & 0.721 & 101 & 0.449 \\
			 \XSolidBold & \CheckmarkBold & \XSolidBold &\CheckmarkBold & 0.300 & 0.540 & 0.718 & 103 & 0.447 \\
			 \XSolidBold & \CheckmarkBold & \CheckmarkBold & \CheckmarkBold & 0.329 & 0.555 & 0.733 & 109 & 0.469 \\
			\CheckmarkBold & \CheckmarkBold & \CheckmarkBold & \CheckmarkBold & \textbf{0.343} & \textbf{0.563} & \textbf{0.752} & 92 & \textbf{0.482} \\
			\bottomrule
	\end{tabular}}

\end{table*}

In this part, we tease apart each module from our standard model and keep other structures as they are, with the goal to justify the contribution of each improvement. The ablation experiments are conducted on the WN18RR dataset, and we find similar performance on all three datasets. Table \ref{tab:test3} shows the results.
After adding a batch-based triple-style negative sampling strategy combined with focal-loss, the appropriate negative sampling ratio dramatically improves the model evaluation effect, as shown in the second line. The original pre-training weights (BERT or RoBERTa pre-trained weights) are not familiar with the corpus information of the link prediction task. After adding the pre-training strategy based on MLM, the evaluation effect is improved. However, the pre-training method based on MLM strategy does not fully excavate the relationship information of triplets, and the multi-task pre-training strategy combined with MLM, MEM, and MRM makes the model evaluation result optimal.

\subsection{Unseen Entities}

To verify the prediction performance of LP-BERT on unseen entities, we re-split the dataset. Specifically, we randomly select 10\% of the entity triples as the validation set and test set, respectively, to ensure that the training set, validation set, and test set don't overlap on any entity. We then re-train and evaluate LP-BERT as well as other baselines, of which the results are shown in Table~\ref{tab:unseen}. 

As can be seen from the table, all models experienced dramatic performance drop across the five metrics. In particular, the distance-based methods are inferior in coping with unseen entities. As mentioned above, such methods only encode the relationship and distance between entities without including semantic information, thus incapable of encoding entities not seen in the training set. Contrastively, pre-training based models, including MLMLM, StAR, and our LP-BERT, have displayed their ability to cope with unseen entities. Furthermore, our LP-BERT surpasses MLMLM and StAR on almost all metrics, proving its superiority for processing unseen entities. Especially, LP-BERT outperforms StAR, the previous state-of-the-art, with over 6-9 points on Hits@3 and Hits@10. However, the score of LP-BERT on the mean rank metric is not as well as other metrics, indicating LP-BERT performs worse on those failed entities. 

\begin{table}[htbp]
	\small
	\centering
	\caption{\label{tab:unseen} Performance of LP-BERT on unseen entities.}
	\setlength{\tabcolsep}{0.7mm}{
		\begin{tabular}{lccccc}
			\toprule
			Models & Hits@1$\uparrow$  & Hits@3$\uparrow$ & Hits@10$\uparrow$ & MRR$\uparrow$ & MR$\downarrow$ \\
			\midrule
			TransE & 0.0010 & 0.0010 & 0.0010 & 0.0010 & 21708 \\
			DistMult & 0.000 & 0.000 & 0.000 & 0.000 & 33955 \\
			ComplEx & 0.000 & 0.000 & 0.000 & 0.000 & 24678 \\
			RotatE & 0.0010 & 0.0010 & 0.0010 & 0.0010 & 21023 \\
			LinearRE & 0.0010 & 0.0010 & 0.0010 & 0.0010 & 21502 \\
			QuatDE & 0.0010 & 0.0010 & 0.0010 & 0.0010 & 21301 \\
			MLMLM & 0.0490 & 0.0932 & 0.1413 & 0.0812 & 6324 \\
			StAR & 0.1108 & 0.2355 & 0.4053 & 0.2072 & \textbf{535} \\
			LP-BERT & \textbf{0.1204} & \textbf{0.2978} & \textbf{0.4919} & \textbf{0.2434} & 1998 \\
			\bottomrule
	\end{tabular}}
\end{table}

\subsection{Case Study}

To further demonstrate the performance of LP-BERT, additional case studies are conducted on the WN18RR datasets and the visualization of the results are provided in Table \ref{tab:case}. In the table, each row denotes a real sample which is randomly selected from the test set.
The first column is a triple which is formatted as (left entity, relation, ?) $\gets$ right entity. The prediction models employ the left entity and the relationship to predict the right entity. From the second column to the fourth column, we present the Top-5 ranked entities with highest predicted probability obtained using different pre-training approaches. The entities are ordered using the predicted probability and the correct answers are highlighted using the bold font. Column 2 presents the predicted results of the proposed pre-training strategy of our proposed LP-BERT. Column 3 provides the results obtained by only using the MLM-based pre-training, while the last column presents the results obtained without pre-training. 

For different approaches, the order of the correctly predicted results is given in the table (the number in each element). For LP-BERT, the orders of the correct predicted results are [1,2,1,2,2], while the orders are [3,3,5,3,3] for MLM-based pre-training and the orders are [6,21,12,6,4] without pre-training. The results suggest that LP-BERT can provide superior performance, with comparison to the MLM-based pre-training the model without pre-training.
Noting that, all the presented results are typical and not cherry-picked for presentation, with the goal to avoid misrepresenting the actual performance of the proposed method.

\begin{table*}[htbp]
		\caption{\label{tab:case} Case study using the WN18RR datasets. The first column is a triple which is formatted as (left entity, relation, ?) $\gets$ right entity. From the second column to the fourth column, we present the Top-5 ranked entities with highest predicted probability which were obtained from different pre-training approaches (including the proposed LP-BERT-based pre-training, MLM-based pre-training and without pre-training). The entities are ordered using the predicted probability. The correct answers are highlighted using the bold font. The number in each element is the order of the correctly predicted results.}
		\centering
		\setlength{\tabcolsep}{1mm}{
			\begin{tabular}{l*{4}{c}}
				\toprule
				\multirow{2}{*}{\textbf {Incomplete Triple}} & \multicolumn{3}{c}{\textbf {Positive entity ranking position \& Top-5 ranked candidate entities}} & \\
				\cmidrule(l){2-4}
				& LP-BERT Pre-training & MLM Pre-training & without Pre-training & \\
				\midrule
				\makecell[c]{(wheeled vehicle, has part, ?)\\ $\gets$ \textbf{axle}} & 
				\makecell[c]{1, (\textbf{axle}, handlebar, wheel spoke,\\hub, splash guard)} &
				\makecell[c]{3, (hub, axletree, \textbf{axle},geared\\ wheel, wheel spoke)} & \makecell[c]{6, (car wheel, bicycle wheel, wagon\\wheel, tyre, axletree)} \\
				\midrule
				\makecell[c]{(heterokontae, hypernym, ?)\\ $\gets$ \textbf{class}} & 
				\makecell[c]{2, (${division^a}$, $\textbf{class}$, kingdom,\\subphylum, ${division^b}$)} &
				\makecell[c]{3, (hub, axletree, \textbf{axle},geared\\ wheel, wheel spoke)} &
				\makecell[c]{21, (protista, algae, euglenophyta,\\pyrrophyta, protoctista)} \\
				\midrule
				\makecell[c]{(chorus line, member meronym, ?)\\ $\gets$ \textbf{showgirl}} & 
				\makecell[c]{1, (\textbf{showgirl}, chorister, chorus line,\\chorus, choir)} &
				\makecell[c]{5, (chorus line, chorus, greek\\chorus, choir, \textbf{showgirl})} &
				\makecell[c]{12, (chorus line, chorus, choir,\\greek chorus, chorus)} \\
				\midrule
				\makecell[c]{(intense, also see, ?)\\ $\gets$ \textbf{profound}} & 
				\makecell[c]{2, (intense, \textbf{profound}, impressive,\\significant, strong)} &
				\makecell[c]{3, (intense, extraordinary,\\\textbf{profound}, wide, large)} &
				\makecell[c]{6, (intense, strong, hot,\\powerful, violent)} \\
				\midrule
				\makecell[c]{(white, synset domain topic of, ?)\\ $\gets$ \textbf{chess game}} & 
				\makecell[c]{2, (board game, \textbf{chess game},\\gameboard, table game, cards)} &
				\makecell[c]{3, ([board game, table game,\\\textbf{chess game}, cards, game)} &
				\makecell[c]{4, (board game, cards, bridge,\\\textbf{chess game}, game)} \\
				
				\bottomrule
		\end{tabular}}
	\end{table*}

\section{Conclusion and Future Work}{\label{conlusion}}

This paper focuses on a fundamental yet critical task in the natural language processing field, i.e., semantic network completion. More specifically, we manage to predict the linkage between entities in the semantic network of the knowledge graph.
We employ the language model and introduce the LP-BERT, which contains multi-task pre-training and knowledge graph fine-tuning phases. In the pre-training phase, we propose two novel pre-training tasks, MEM and MRM, to encourage the model to learn the knowledge of context and the structure information of the knowledge graph.
While in the fine-tuning phase, we design a triple-style negative sampling in a batch, which greatly increases the proportion of negative sampling while keeping the training time almost unchanged. Extensive experimental results on three datasets demonstrated the efficiency and effectiveness of our proposed LP-BERT.
In future work, we will explore more diverse pre-training tasks and increase the model parameter size to enable LP-BERT to store larger graph knowledge.

\section{Acknowledgements}
This work is partially supported by the National Key Research and Development Program of China (2021ZD0112901).

\bibliographystyle{IEEEtran.bst}
\bibliography{ref.bib}


%

%
%
%
%
%


\ifCLASSOPTIONcaptionsoff
  \newpage
\fi

\end{document}